\pdfoutput=1

\documentclass[10pt,twocolumn,letterpaper]{article}

\usepackage{cvpr}       

\usepackage{graphicx}
\usepackage{amsmath}
\usepackage{amssymb}
\usepackage{booktabs}

\usepackage[pagebackref,breaklinks,colorlinks]{hyperref}

\usepackage[capitalize]{cleveref}
\crefname{section}{Sec.}{Secs.}
\Crefname{section}{Section}{Sections}
\Crefname{table}{Table}{Tables}
\crefname{table}{Tab.}{Tabs.}

\def\cvprPaperID{*****} 
\def\confName{CVPR}
\def\confYear{2022}

\begin{document}

\title{Frame-level Prediction of Facial Expressions, Valence, Arousal and Action Units for Mobile Devices}

\author{Andrey V. Savchenko\\
HSE University\\
Laboratory of Algorithms and Technologies for Network Analysis, Nizhny Novgorod, Russia\\
{\tt\small avsavchenko@hse.ru}
}
\maketitle

\begin{abstract}
  In this paper, we consider the problem of real-time video-based facial emotion analytics, namely, facial expression recognition, prediction of valence and arousal and detection of action unit points. We propose the novel frame-level emotion recognition algorithm by extracting facial features with the single EfficientNet model pre-trained on AffectNet. As a result, our approach may be implemented even for video analytics on mobile devices. Experimental results for the large scale
Aff-Wild2 database from the third Affective Behavior Analysis in-the-wild (ABAW) Competition demonstrate that our simple model is significantly better when compared to the VggFace baseline. In particular, our method is characterized by 0.15-0.2 higher performance measures for validation sets in uni-task Expression Classification, Valence-Arousal Estimation and Expression Classification. Due to simplicity, our approach may be considered as a new baseline for all four sub-challenges.
  \end{abstract}

\section{Introduction}
\label{sec:intro}
Affective computing, emotional intelligence~\cite{pietikainen2022challenges} and, in particular, analysis of humans' emotional states based on facial videos, is an essential task for many systems with man-machine interaction~\cite{deng2021iterative}, education, health~\cite{thinh2021emotion} and mobile services~\cite{demochkina2021mobileemotiface,grechikhin2019user,savchenko2021user}. Many facial analysis tasks, such as face recognition, age and gender prediction, have reached high accuracy appropriate for many practical applications~\cite{cao2018vggface2,savchenko2019peerj}. However, but the ability to understand human emotions is still far from maturity~\cite{jin2021multi}. The personal bias and backgrounds increase the
uncertainty of emotion perception and contextual information~\cite{deng2021iterative}. As a result, the datasets used to train FER (facial expression recognition) models are not very large and contain a lot of noise and inconsistencies in emotional labels of photos~\cite{mollahosseini2017affectnet}. The video-based FER is even more complex task, because human emotions may change rapidly, and many frames do not contain enough information to reliably predict facial (macro) expression. Hence, the authors of the datasets are required to provide the labeling at frame level~\cite{saeed2014frame}. Thus, the number of video datasets for in-the-wild affective computing has been very limited.

The situation has changed with an appearance of the AffWild dataset~\cite{kollias2019deep,zafeiriou2017aff}. It has been recently extended in the AffWild2 database~\cite{kollias2019expression} with more videos and annotations for the following tasks: (1) frame-level FER; (2) detection of action units (AU), i.e., specific movements of facial muscles from Facial Action Coding System (FACS)~\cite{ekman2005face}; and (3) prediction of valence and arousal, i.e., how active or passive, positive or negative is the human behavior. 

Though FER has been a topic of major research~\cite{kollias2021affect}, many models learn too many features specific for a concrete dataset, which is not practical for in-the-wild settings~\cite{jin2021multi}. The development of in-the-wild affect prediction engines has been accelerated by a couple of ABAW (Affective Behavior Analysis in-the-wild) competitions~\cite{kollias2020analysing,kollias2021analysing}. The third place in the first and second tasks was achieved by the authors of the paper~\cite{thinh2021emotion} who proposed the multi-task learning (MTL) technique for
the incomplete labels of these correlated tasks. The multi-modal audiovisual ensemble model~\cite{jin2021multi} took the second place, while the winner of these two sub-challenges was a multi-task streaming network~\cite{zhang2021prior}. The latter captures identity-invariant emotional features using an advanced facial embedding. The valence-arousal challenge was won by deep ensembles with iterative distillation and pseudo-labeling~\cite{deng2021iterative}.

As one can notice, most successful previous solutions use MTL~\cite{kollias2021distribution,zhang2021prior} to boost their performance. As a result, the authors of the third ABAW contest~\cite{kollias2022abaw} decided to inspire researchers studying not only MTL, but also the uni-task models. The baseline uses the deep convolutional neural network (CNN), namely, VGG16, pre-trained on the VGGFACE dataset to make a decision in all tasks independently~\cite{kollias2022abaw}. 

In this paper, we discuss our solution for all four tasks from the ABAW3 challenge. Most participants of such contests are mainly focused on improvement of accuracy metrics, so that they usually develop complex ensemble models~\cite{deng2021iterative,zhang2021prior}. Hence, they cannot be implemented in real-time analysis of affective behavior in mobile or embedded systems. Hence, the main motivation of this paper is to develop a single~\cite{kollias2019face} and lightweight model~\cite{savchenko2021emotions} that not only achieve high accuracy but may be used in mobile applications~\cite{grechikhin2019user}. As a result, we contributed the novel model based on the EfficientNet architecture~\cite{tan2019efficientnet} that is much better than the baseline in terms of both size and performance. The weights of our CNN are tuned on external AffectNet dataset~\cite{mollahosseini2017affectnet}, so the facial embeddings extracted by this neural network do not learn any features that are specific to the Aff-Wild2 dataset. Thus, our method may become a new baseline for future challenges with the ABAW challenges.
 
This paper is structured as follows. Section~\ref{sec:2} introduces our efficient model and the training procedure. Experimental results for all tasks of ABAW challenge are presented in Section~\ref{sec:3}. Finally, concluding comments and future work are discussed in Section~\ref{sec:4}.
 
\section{Proposed approach}\label{sec:2}

Three frame-level affective behavior analysis tasks~\cite{kollias2022abaw} are considered for an input video $\{X(t)\}, t=1,2,...,T$ with $T$ frames, namely:
\begin{enumerate}
 \item Facial expression recognition, in which it is required to assign each frame $X(t)$ to one of $C_{EXPR}>1$ categories (classes), such as happiness, fear, etc. It is a general multi-class classification problem.
 \item AU analysis and recognition, in which the frame $X(t)$ is associated with a subset of $C_{AU} \ge 1$ AUs. The task may be considered as a multi-label classification problem, i.e., prediction of a binary vector $\mathbf{AU}(t)=[AU_1(t),...,AU_{C_{AU}}(t)]$. Here, $AU_i(t)=1$ if the $i$-th action unit is detected in the $t$-th frame, otherwise $AU_i(t)=0$.
 \item Prediction of valence and arousal of an emotion. It is a type of regression tasks, but the values of valence and arousal are typically limited to a range $[-1,1]$.
\end{enumerate}
 
It is assumed that the facial regions are preliminarily extracted, so that $X(t)$ contain cropped faces. The supervised learning scenario is considered, where a training set of $N$ reference facial images $\{X_n\}, n \in \{1,...,N\}$ are given, and the facial expression $e_n\in \{1,...,C_{EXPR}\}$, $C_{AU}$-dimensional binary vector $\mathbf{AU}_n$ of action units, valence $V_n$, arousal $A_n$ of the $n$-th reference image are known, though it is possible that some labels are unavailable. 

\begin{figure}[t]
 \centering
 \includegraphics[width=0.95\linewidth]{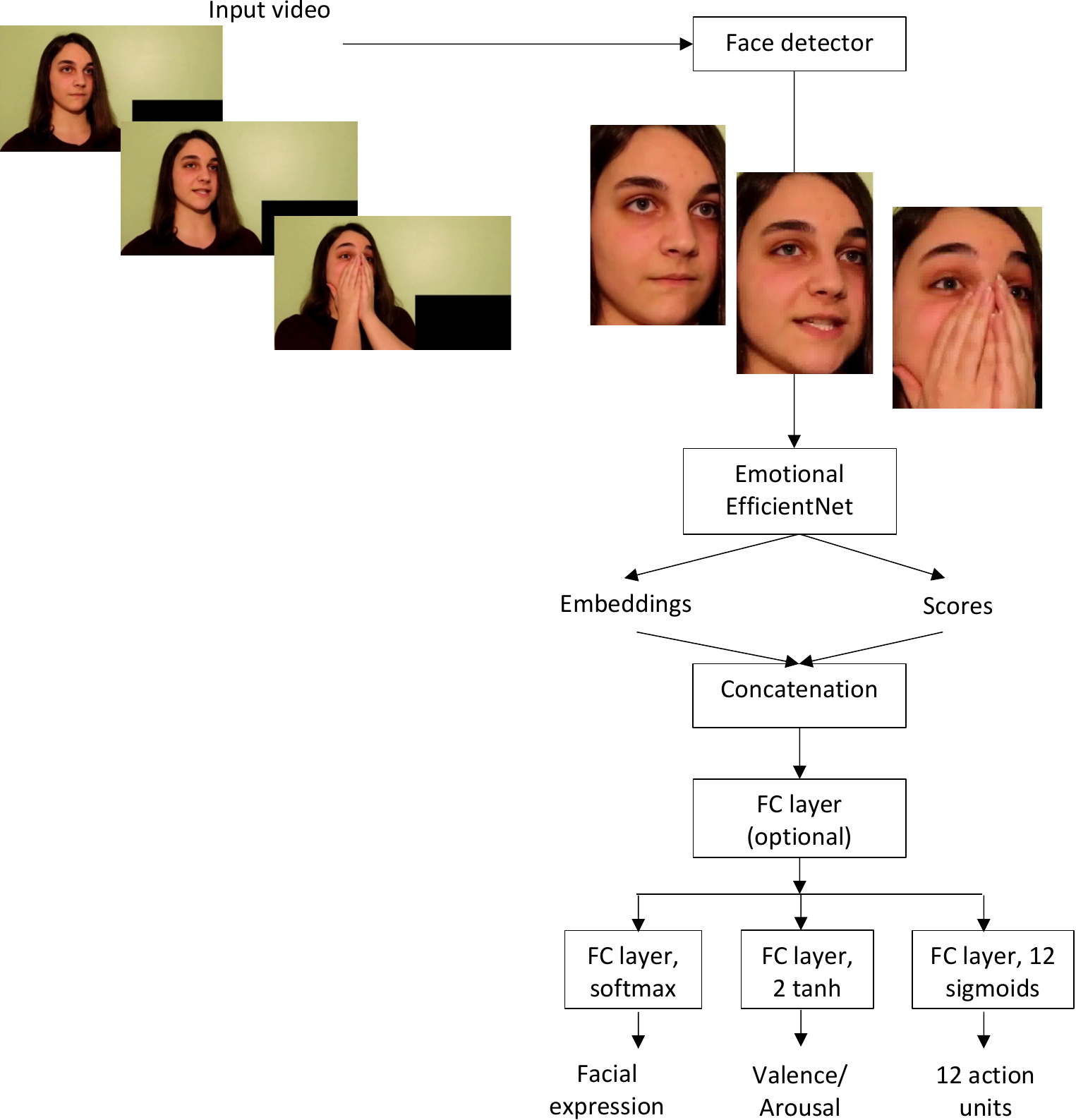}
 \caption{Proposed model for the multi-task learning.}
 \label{fig:1}
\end{figure}

In this paper, we will use conventional approach based on pre-training of deep CNN using large FER dataset. As we pay special attention to offline recognition on mobile devices~\cite{savchenko2022preference}, it is reasonable to use such architectures as MobileNets or EfficientNets~\cite{tan2019efficientnet}. The presented approach consists of the following steps:
\begin{enumerate}
 \item Pre-training of a lightweight model on face identification task using very large facial dataset of celebrities~\cite{cao2018vggface2}.
 \item Fine-tuning the model from item (1) on static photos from external dataset to obtain an emotional CNN~\cite{mollahosseini2017affectnet}.
 \item The outputs of the emotional CNN (embeddings and expression scores) from item (2) are used to extract facial features of each video frame from the AffWild2 dataset~\cite{kollias2019expression}.
 \item These embeddinsg and scores are used to train simple frame-level MLP-based classification/regression models given the training set of each challange.
 \item Optional post-processing of frame-level outputs on models from item (4) computed for validation and test sets to make the predictions more smooth.
\end{enumerate}

Let us consider the details of our approach. At first, a large external VGGFace2 facial dataset~\cite{cao2018vggface2} with 9131 subjects is used to pre-train a CNN on face recognition task. The faces cropped by MTCNN (multi-task cascaded neural network)~\cite{zhang2016joint} detector without any margins were utilized for training, so that most parts of the background, hairs, etc. is not presented. As a result, the learned facial features are more suitable for emotional analysis. We trained the model totally of 8 epochs by the Adam optimizer and SAM (Sharpness-Aware Minimization)~\cite{foret2020sharpness}. The models with the highest accuracy on validation set, namely, 92.1\%, 94.19\% and 95.49\% for MobileNet-v1, EfficientNet-B0 and EfficientNet-B2, respectively, were used.

Second, the resulted CNN is fine-tuned on the training set of 287,651 photos from the AffectNet dataset~\cite{mollahosseini2017affectnet} annotated with $C=8$ basic expressions (Anger, Contempt, Disgust, Fear, Happiness, Neutral, Sadness and Surprise). It is necessary to emphasize that the annotations of valence and arousal from the AffectNet dataset were not used in the pre-training. The last layer of the network pre-trained on VGGFace2 is replaced by the new head (fully-connected layer with $C$ outputs and softmax activation), so that the penultimate layer with $D$ neurons can be considered as an extractor of facial features. The weighted categorical cross-entropy (softmax) loss was optimized~\cite{mollahosseini2017affectnet}. The new head was trained during 3 epochs with learning rate 0.001. Finally, the whole network is fine-tuned with a learning rate of 0.0001 at the last 5 epochs. The details of this training procedure are available in~\cite{savchenko2021emotions}. As a result, we fine-tuned three models, namely, MobileNet-v1, EfficientNet-B0 and EfficientNet-B2, that reached accuracy 60.71\%, 61.32\% and 63.03\%, on the validation part of the AffectNet.

Third, such an emotional CNN was used as a feature extractor for frames $X(t)$ and reference images $X_n$. Though the cropped facial images provided by the organizers of the challenge have different (typically, low) resolution, they were resized to 224x224 pixels for the first two models, while the latter CNN requires input images with resolution 300x300. We examine two types of features: (1) facial image embeddings (output of penultimate layer)~\cite{savchenko2021emotions,tseytlin2021hotel}; and (2) scores (predictions of emotional class probabilities at the output of last softmax layer). As a result, $D$-dimensional embeddings $\mathbf{x}(t)$ and $\mathbf{x}_n$ and $C$-dimensional scores $\mathbf{s}(t)$ and $\mathbf{s}_n$ are obtained. Three kinds of features have been examined, namely: (1) embeddings only; (2) scores only; and (3) concatenation of embeddings and scores~\cite{rassadin2017group}. According to the rules of the uni-task challenges, the pre-trained model can be pre-trained on any task (e.g., VA estimation, Expression Classification, AU detection, Face Recognition), so that the expression scores returned by our model trained on the AffectNet can be used as facial features to predict Valence/Arousal and AUs. When we refined the model given the ABAW3 dataset, only the annotations available for a concrete challenge have been used to train a classification and regression models. 

\begin{figure}[t]
 \centering
 \includegraphics[width=0.95\linewidth]{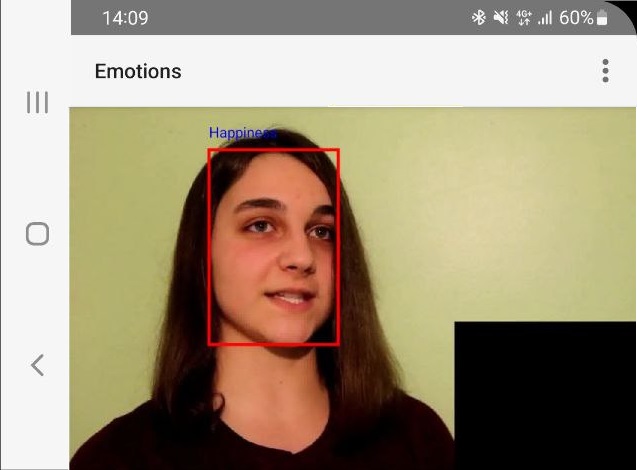}
 \caption{Sample screen of Android demo application.}
 \label{fig:2}
\end{figure}

Fourth, we trained a shallow feed-forward neural network, such as multi-class logistic regression or MLP (multi-layered perceptron) with one hidden layer for each of three tasks as follows:
\begin{enumerate}
 \item The output layer for expression recognition task contains $C_{EXPR}$ neurons with softmax activation. The weighted categorical cross-entropy was minimized for the first task. The final solution is taken in favor of facial expression with the maximal predicted probability.
 \item Two neurons with $tanh$ activations are used at the last layer to predict valence and arousal. The loss function is computed as $1-0.5(CCC_V+CCC_A)$~\cite{kollias2021affect}, where $CCC_V$ and $CCC_A$ are estimates of the Concordance Correlation Coefficient (CCC) for valence and arousal, respectively.
 \item Action unit detector contains $C_{AU}$ output units with sigmoid activation. The weighted binary cross-entropy loss is minimized. To predict the final binary vector, the outputs of this model are matched with a fixed threshold. We examine two possibilities, namely, one threshold (0.5) for each action unit or individual threshold for each action unit. In the latter case, the best threshold is chosen from the list of 10 values $\{0.1, 0.2,..., 0.9\}$ by maximizing the class-level F1 score for the validation set. 
\end{enumerate}

\begin{table*}
 \centering
 \begin{tabular}{cccc}
 \toprule
 Model & Method & F1-score $P_{EXPR}$ & Accuracy \\
 \midrule
 VGGFACE & Baseline~\cite{kollias2022abaw} & 0.23 & -\\
 \hline
	 & embeddings & 0.285 & 0.398 \\
 Our&embeddings, 1 hidden layer & 0.338 & 0.460 \\
	MobileNet~\cite{demochkina2021mobileemotiface} & scores & 0.236 & 0.435 \\
 &scores, 1 hidden layer & 0.286 & 0.433 \\
 \hline
	Our EfficientNet-& embeddings & 0.307 & 0.428 \\
 B0~\cite{savchenko2021emotions}&embeddings, 1 hidden layer & \bf 0.381 & \bf 0.500 \\
 \hline
	Our EfficientNet-& embeddings & 0.305 & 0.412 \\
 B2&embeddings, 1 hidden layer & 0.317 & 0.435 \\
 \bottomrule
 \end{tabular}
 \caption{Expression Challenge Results on the Aff-Wild2’s validation set.}
 \label{tab:expr}
\end{table*}

The model for each task is trained on 20 epochs with early stopping and Adam optimizer (learning rate 0.001). Fig.~\ref{fig:1} contains the most general case of the proposed model with three outputs is trained for the multi-task learning challenge. If the uni-task challenge is considered, only one output layer is used. Here, the facial regions are detected in each frame using MTCNN. The emotional features are extracted using our EfficientNet model. These features are fed into MLP to solve one of the tasks or all tasks together in the multi-task learning scenario. If the facial region is not detected in a couple of frames, we perform the bilinear interpolation of the outputs of the model for two frames with detected faces. If face detection fails for several first or last frames of the video, we will simply use predictions for the closest frame with detected face. 

Fifth, it is possible to smooth the predictions for $k \ge 1$ consecutive frames by using point-wise mean (box) or median filter with kernel size $k$~\cite{rassadin2017group}. If $k$ is equal to 1, the frame-level predictions will be used. Otherwise, the slicing window with size $k$ is processed for every $t$-th frame, i.e., we took the predictions at the output of our MLP classifiers for frames $t-\frac{k}{2}, t-\frac{k}{2}+1,...,t-1,t+1,...,t+\frac{k}{2}-1,t+\frac{k}{2}$. The final decision function for the frame $t$ is computed as a point-wise mean or median of these $k$ predictions.

The training script for the presented approach is made publicly available\footnote{\url{https://github.com/HSE-asavchenko/face-emotion-recognition/blob/main/src/abaw_train.ipynb}}. The CNNs used for feature extraction, namely, MobileNet v1 (TensorFlow's mobilenet\_7.h5) and EfficientNets (PyTorch's enet\_b0\_8\_best\_vgaf and enet\_b2\_8), are also available in this repository\footnote{\url{https://github.com/HSE-asavchenko/face-emotion-recognition/tree/main/models/affectnet_emotions}}. Finally, the possibility to use our model for mobile devices is demonstrated. The sample output of the demo Android application is presented in Fig.~\ref{fig:2}. It is possible to recognize facial expressions of all subjects in either any photo from the gallery or the video captured from the frontal camera. 

\section{Experimental results}\label{sec:3}

In this section, four tasks from the third ABAW challenge are considered. We used the cropped images officially provided by the organizers of this challenge. 

\begin{table}
 \centering
 \begin{tabular}{@{}lc@{}}
 \toprule
 Expression & F1-score \\
 \midrule
 Neutral & 0.609\\
 Anger & 0.151\\
 Disgust & 0.516\\
 Fear & 0.016\\
 Happiness & 0.477\\
 Sadness & 0.461\\
 Surprise & 0.303\\
 Other & 0.512\\
 \bottomrule
 \end{tabular}
 \caption{F1 scores for FER with the best EfficientNet-B0 model.}
 \label{tab:expr_f1}
\end{table}

\subsection{Uni-task Expression Recognition}
The first experiment is devoted to the uni-task FER with $C_{EXPR}=8$ classes (anger, disgust, fear, happiness, sadness, surprise, neutral and other). The frame files missed in the cropped directory were ignored. As a result, the training and validation sets contains 585,317 and 280,532 files, respectively. Two performance metrics were computed, namely, (1) macro-averaged F1 score $P_{EXPR}$~\cite{kollias2022abaw}; and (2) top-1 unbalanced accuracy. The ablation study for several feature extractors and classifiers is presented in Table~\ref{tab:expr}. Here, the absence of prefix ``1 hidden layer" stands for the neural network without hidden layers. First, embeddings are classified more accurately when compared to emotional scores. Second, though EfficientNet-B2 has 2\% greater accuracy than EfficientNet-B0 on the validation part of AffectNet~\cite{savchenko2021emotions}, the latter model provides much better performance in this challenge. As a result, our best mean F1-score is 0.16 higher than $P_{EXPR}$ of the baseline VGGFACE provided by organizers of this challenge. However, even we used the weighted cross-entropy as a loss function, the imbalance of the dataset still influences the overall quality. Table~\ref{tab:expr_f1} demonstrates the F1 scores of our best model for each class. As one can notice, the quality for anger and, especially, fear emotions is very low.

\begin{table*}
 \centering
 \begin{tabular}{cccc}
 \toprule
 Model & Method & F1-score, threshold 0.5 & F1-score $P_{AU}$, different thresholds \\
 \midrule
 VGGFACE & Baseline~\cite{kollias2022abaw} & - & 0.39\\
 \hline
	& embeddings & 0.473 & 0.524 \\
 Our &embeddings, 1 hidden layer & 0.477 & 0.529 \\
	 MobileNet~\cite{demochkina2021mobileemotiface} & scores & 0.432 & 0.442 \\ 
 &scores, 1 hidden layer & 0.452 & 0.487 \\
 \hline
	Our EfficientNet-& embeddings & 0.491 & 0.518 \\
 B0~\cite{savchenko2021emotions}&embeddings, 1 hidden layer & \bf 0.508 & \bf 0.537 \\ 
 \hline
	Our EfficientNet-& embeddings & 0.468 & 0.503 \\
 B2&embeddings, 1 hidden layer & 0.482 & 0.512 \\
 \bottomrule
 \end{tabular}
 \caption{Action Unit Challenge Results on the Aff-Wild2’s validation set.}
 \label{tab:au}
\end{table*}

\begin{table*}
 \centering
 \begin{tabular}{ccccc}
 \toprule
 Model & Method & CCC\_V & CCC\_A& Mean CCC $P_{VA}$ \\
 \midrule
 ResNet-50 & Baseline~\cite{kollias2022abaw} & 0.31 & 0.17 & 0.24\\
 \hline
 Our &embeddings &0.303 & 0.449 & 0.376 \\
	 MobileNet~\cite{demochkina2021mobileemotiface}& scores & 0.404 & 0.423 & 0.413 \\
 \hline
	Our EfficientNet-& embeddings & 0.309 & 0.436 & 0.372 \\
 B0~\cite{savchenko2021emotions}&scores& \bf 0.429 & \bf 0.496 & \bf 0.463 \\
 \hline
	Our EfficientNet-& embeddings & 0.377 & 0.474 & 0.426 \\
 B2&scores & 0.408 & 0.477 & 0.443 \\
 \bottomrule
 \end{tabular}
 \caption{Valence-Arousal Challenge Results on the Aff-Wild2’s validation set.}
 \label{tab:va}
\end{table*}

\begin{table*}
 \centering
 \begin{tabular}{cc|cccc|cccc}
 \toprule
 & &\multicolumn{4}{c|}{Validation set} & \multicolumn{4}{c}{Test set} \\
 Model & Smoothing & $P_{EXPR}$ & $P_{AU}$ & CCC\_V & CCC\_A & $P_{EXPR}$ & $P_{AU}$ & CCC\_V & CCC\_A\\
 \midrule
Baseline~\cite{kollias2022abaw}& - & 0.23 & 0.39 & 0.31 & 0.17 & 0.205 & 0.365 & 0.18 & 0.17 \\
 \hline
	& Frame-level ($k=0$) & 0.3807 & 0.5367 & 0.4297 & 0.4980 & 0.2926 & 0.4660 & 0.4014 & 0.4278\\
	Proposed & Mean ($k=5$) & 0.3914 & 0.5447 & 0.4375 & 0.5132 & 0.2973 & 0.4718 & 0.4083 & 0.4389\\
	model& Median ($k=5$) & 0.3888 & 0.5430 & 0.4354 & 0.5072 & 0.2964 & 0.4705 & 0.4061 & 0.4349\\
	& Mean ($k=15$) & \bf 0.4018 & 0.5445 & \bf 0.4485 & \bf 0.5353 & \bf 0.3025 & 0.4713 & \bf 0.4174 & \bf 0.4538\\
	& Median ($k=15$) & 0.3996 & \bf 0.5478 & 0.4459 & 0.5272 & 0.3007 & \bf 0.4731 & 0.4135 & 0.4488\\
 \bottomrule
 \end{tabular}
 \caption{Results of smoothing techniques on the Aff-Wild2’s validation and test sets.}
 \label{tab:smooth}
\end{table*}

\begin{table*}
 \centering
 \begin{tabular}{cccccc}
 \toprule
 Model & Method & $P_{MTL}$ & $P_{EXPR}$ & $P_{VA}$ & $P_{AU}$ \\
 \midrule
 VGGFACE & Baseline~\cite{kollias2022abaw}& 0.30 & - & - & - \\
 \hline
	Our & embeddings + scores & 1.112 & 0.358 & 0.282 & 0.471 \\
 MobileNet~\cite{demochkina2021mobileemotiface}&embeddings + scores, 1 hidden layer & 1.037 & 0.321 & 0.252 & 0.464\\
 \hline
	Our & embeddings + scores & 1.123 & \bf 0.386 & 0.283 & 0.455 \\
 EfficientNet-B0~\cite{savchenko2021emotions}&embeddings + scores, 1 hidden layer & 1.121 & 0.381 & 0.272 & 0.469 \\
 \hline
	& embeddings + scores & 1.147 & 0.384 & \bf 0.302 & 0.461 \\
 Our EfficientNet-B2&embeddings + scores, 1 hidden layer & 1.135 & 0.378 & 0.298 & 0.458 \\
 \cline{2-6}
 	& embeddings + scores, different AU thresholds& \bf 1.150 & 0.384 & \bf 0.302 & \bf 0.490 \\
 \bottomrule
 \end{tabular}
 \caption{Multi-Task-Learning Challenge Results on the Aff-Wild2’s validation set.}
 \label{tab:mtl}
\end{table*}

\subsection{Uni-task Action Unit Detection}

In the second experiment, we examine the uni-task Action Unit Detection problem. The training set contains 1,356,861 images and $C_{AU}=12$ action units (AU1, AU2, AU4, AU6, AU7, AU10, AU12, AU15, AU23, AU24, AU25 and
AU26), while 445836 facial frames are included into the validation set. The unweighted average F1-score $P_{AU}$ of our models (Table~\ref{tab:au}) is again up to 0.14 points greater than the baseline performance. This table contains the results when the threshold for each AU is equal to 0.5, and for different thresholds tuned for each AU separately. In the latter case, the following 12 thresholds were automically found using the validation set: 0.8, 0.8, 0.7, 0.5, 0.5, 0.5, 0.6, 0.8, 0.8, 0.8, 0.3, and 0.7. The results are very similar to the first experiment: embeddings are classified more accurately, and EfficientNet-B0 is the best model. 

\subsection{Uni-task Valence-Arousal Prediction}

In the third experiment, the uni-task Valence-Arousal Estimation is analyzed. The number of labeled images here is much higher, so that 1,555,919 and 338,755 frames were put into the training and validation sets. The estimates of CCC for valence and arousal together with their mean $P_{VA}$ are shown in Table~\ref{tab:va}. As one can notice, EfficientNet-B0 is still the best model, which has twice-higher $P_{VA}$ when compared to the baseline~\cite{kollias2022abaw}. In contrast to the previous experiment, the greatest CCC is achieved by very fast regression models for $C=8$ emotional scores of the AffectNet dataset. Conventional usage of embeddings leads to 0.02-0.09 lower CCCs.

\subsection{Aggregation of Frame-Level Predictions}

Next, we studied the impact of smoothing on the performance measures for EfficientNet-B0 and the best classifiers from the previous experiment. Five submissions have been prepared for AU, EXPR and VA challenges by using the frame-level predictions without smoothing and box (mean) and median filters with kernel sizes $k=5$ and $k=15$. The performance measures on the validation and test sets for each challenge are shown in Table~\ref{tab:smooth}. The best results are obtained by using the large slicing window ($k=15$ frames). The mean filter is in most cases better than the median filter except the AU challenge. The proposed approach is much more accurate than the baseline on both validation and test sets. For example, our model has 10\% greater F1-scores for the test set from the Expression and Action Unit Challenges. The most impressive is the increase of the CCC metric in the Valence-Arousal Challenge where we improved the baseline by more than 20\%.

\subsection{Multi-Task Learning}

In the last experiment, we trained a complete multi-task model (Fig.~\ref{fig:1}) on the set of 142,225 images and computed the metric $P_{MTL}=P_{EXPR}+P_{VA}+P_{AU}$ using 26,876 validation frames. The results are reported in Table~\ref{tab:mtl}. It is important to notice two differences with previous experiments: (1) the best model here is EfficientNet-B2, and (2) MLP with 1 hidden layer is worse than a very simple logistic regression. In fact, the latter does not use knowledge about multiple tasks and makes predictions for each task independently. During the challenge, we made four submissions by using two CNNs (EfficientNet-B0 and B2) and training the feed-forward neural network classifier on the training set and concatenation of the training and validation sets. Despite the superiority of EfficientNet-B2 on the validation set (Table~\ref{tab:mtl}), the best performance metric $P_{MTL}=0.809$ is obtained for EfficientNet-B0, though the difference with EfficientNet-B2 ($P_{MTL}=0.8083$) is not significant. Anyway, our simple model is much more accurate when compared to the baseline. Indeed, its metric $P_{MTL}$ is equal to 0.3 and 0.28 for validation and test set, respectively. Thus, we improved the performance by more than 50\% and took the third place in MTL sub-challenge. 

\section{Conclusion}\label{sec:4}

In this paper, we have presented the frame-level facial emotion analysis model (Fig.~\ref{fig:1}) based on concatenation of embeddings and scores at the output of EfficientNet. The latter CNN was carefully pre-trained on the VGGFace2~\cite{cao2018vggface2} and AffectNet~\cite{savchenko2021emotions} datasets. Its experimental study for the tasks of the third ABAW challenge~\cite{kollias2022abaw} have demonstrated that our technique is much more accurate than the baseline VGGFACE. Moreover, we have implemented an Android mobile application (Fig.~\ref{fig:2}) with publicly available source code to demonstrate the real-time efficiency of our approach and motivate practitioners to implement the facial emotion analytic engines on-device. Our approach is based on a single lightweight neural network, so that it may be not as accurate as ensembles of many CNNs~\cite{deng2021iterative}.

Our best EfficientNet-B0 model is characterized by F1-score 0.38 for expression recognition, mean CCC 0.46 for valence and arousal prediction, and F1-score 0.54 for action unit detection. Our approach took the third place in the MTL challenge, fourth places in the Expression and Valence-Arousal Challenges and fifth place in the Action Unit Challenges. In average, there is only one team who took slightly lower average place in all four tasks~\cite{zhang2022transformer}. Their transformer-based multimodal solution is the winner in the uni-task Expression Classification and Action Unit Detection, but our method is better in other two challenges. As the proposed model has not been fine-tuned on the AffWild2 dataset, we can claim that the facial features extracted by our networks lead to the most robust decisions. Due to simplicity, our approach may be considered as a new baseline for all four sub-challenges.

In the future, it is necessary to integrate our approach into more complex pipelines. For example, we process frames independently, so that any sequential and attention models can benefit from the usage of our facial emotional features~\cite{demochkina2021neural,makarov2022self,meng2019frame}. Moreover, it is worth studying the combination of our models into a large ensemble with different representations of input videos. Finally, as our current best model for MTL solves each task independently, it is important to properly use the correlation between different tasks in the multi-task learning scenario~\cite{zhang2021prior}.

\textbf{Acknowledgements}. The work is supported by RSF (Russian Science Foundation) grant 20-71-10010.

{\small
\bibliographystyle{ieee_fullname}
\bibliography{paper}
}

\end{document}